\pdfoutput=1

\documentclass[11pt]{article}

\usepackage[]{acl}

\usepackage{times}
\usepackage{latexsym}
\usepackage{multirow}
\usepackage{hyperref}

\usepackage{amssymb}
\usepackage{amsmath}
\usepackage{marvosym}
\usepackage{graphicx}
\newcommand{\model}{ PASA}

\usepackage[T1]{fontenc}

\usepackage[utf8]{inputenc}

\usepackage{microtype}

\title{Learn What Is \textit{Possible}, Then Choose What Is \textit{Best}: Disentangling One-To-Many Relations in Language Through Text-based Games}

\author{Benjamin Towle\textsuperscript{1}, Ke Zhou\textsuperscript{1,2} \\
  \textsuperscript{1}University of Nottingham \\
  \textsuperscript{2}Nokia Bell Labs \\
  \texttt{\{benjamin.towle, ke.zhou\}@nottingham.ac.uk} \\\
}

\begin{document}
\maketitle
\begin{abstract}
Language models pre-trained on large self-supervised corpora, followed by task-specific fine-tuning has become the dominant paradigm in NLP. These pre-training datasets often have a one-to-many structure---e.g. in dialogue there are many valid responses for a given context. However, only some of these responses will be desirable in our downstream task.  This raises the question of how we should train the model such that it can emulate the desirable behaviours, but not the undesirable ones. Current approaches train in a one-to-one setup---only a single target response is given for a single dialogue context---leading to models only learning to predict the \textit{average} response, while ignoring the full range of \textit{possible} responses. Using text-based games as a testbed, our approach, PASA, uses discrete latent variables to capture the range of different behaviours represented in our larger pre-training dataset. We then use knowledge distillation to distil the posterior probability distribution into a student model. This probability distribution is far richer than learning from only the hard targets of the dataset, and thus allows the student model to benefit from the richer range of actions the teacher model has learned. Results show up to 49\% empirical improvement over the previous state-of-the-art model on the Jericho Walkthroughs dataset \footnote{Paper was published in EMNLP Findings 2022. Available at \url{https://aclanthology.org/2022.findings-emnlp.364/}}.
\end{abstract}

\section{Introduction}

\begin{figure*}
    \centering
    \includegraphics[scale=0.105]{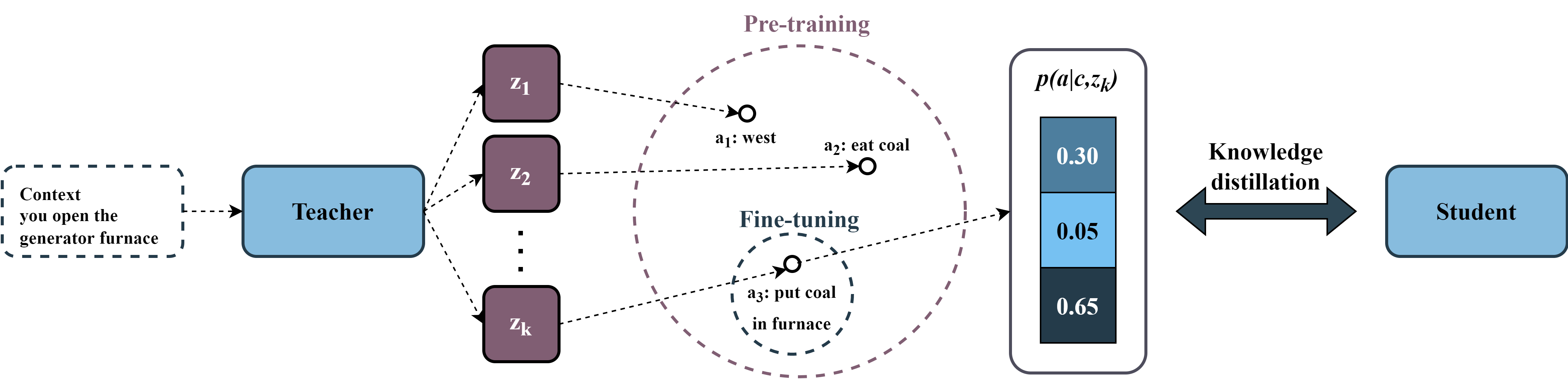}
    \caption{Overview for PASA. A teacher model learns to predict $K$ different actions to capture the pre-training data distribution, and the posterior probability from the fine-tuning data is distilled into a student model.}
    \label{fig:overview}
\end{figure*}

Language models pre-trained on large self-supervised corpora, followed by downstream fine-tuning on the target domain has become a popular recipe in NLP \citep{roller-etal-2021-recipes, Wolf2019TransferTransfoAT}. These pre-training datasets often have a one-to-many structure---e.g. in dialogue there are many valid responses for a given context. However, only some of these responses will be desirable in our downstream task, while others will be counterproductive. For instance, dialogue models intended to engage the user in friendly chitchat are pre-trained on Reddit \cite{dialogpt}, which contains a mixture of users who are thoughtful and interesting as well as hostile and argumentative \citep{chatbotsafety}. This raises the question of how we should train the model such that it can emulate the desirable behaviours on the downstream task, but not the undesirable ones.

The current predominant approach is to train in a one-to-one setup---only a single target response is given for a single dialogue context \citep{roller-etal-2021-recipes, Wolf2019TransferTransfoAT}. However, this leads to models only learning to predict the \textit{average} response, while ignoring the full range of \textit{possible} responses \citep{li-etal-2016-diversity}. This is particularly problematic when the desired responses are only contained in a small subset of the dataset.

To explore solutions we leverage the domain of text-based games \citep{colossal}. Text-based games involve interacting in a textual world by issuing commands ("actions") in natural language---which are interpreted by the game's hard-coded parser---and receiving textual observations from the game's interface (see Figure \ref{fig:overview}). The objective is to maximise the integer score by completing various in-game tasks such as exploring new areas, collecting valuable items or solving puzzles. Text-based games are especially well-suited to the task for two key reasons: 
\begin{itemize}
    \item There are two relevant datasets---ClubFloyd \citep{yao2020calm} and Jericho Walkthroughs \citep{colossal}---which mirror the aforementioned situation in dialogue. ClubFloyd is a comparatively large dataset of humans playing text-based games, and therefore contains a mixture of behaviours, ranging from expert humans who obtain the maximum score, to novice humans who obtain virtually no score, which parallels the mixture of friendly and unfriendly dialogue behaviours in the aforementioned Reddit example. Jericho Walkthroughs is a much smaller dataset containing only the exact actions required to obtain the maximum score. Thus, ClubFloyd represents our pre-training dataset containing a mixture of desirable and undesirable behaviour and Jericho Walkthroughs represents our fine-tuning dataset containing only the desirable behaviour. 
    \item Through backend access to the game engine, we can train and evaluate directly with the set of valid actions---those actions that are understood by the game's parser given the current game state. Unlike dialogue, where most datasets do not contain the set of valid responses, this allows us to explicitly evaluate the task of predicting the best (sc. \textit{gold}) action---the action that maximises long-term score---given the set of valid actions, which forms the key focus for this paper.
\end{itemize}

Our proposed approach, PASA (\textbf{P}osterior \textbf{A}ction \textbf{S}election from Discrete L\textbf{a}tent Variables), uses discrete latent variables to capture the range of different behaviours represented in our larger pre-training dataset. Particularly, we learn to predict a set of possible actions, where each action reflects a different value for the latent variable. We assign our latent variable using the ground-truth action itself, via a posterior recognition network. Unlike previous approaches, we do not then learn to predict the latent variable from the context. This is because our goal is not to learn to output diverse responses as is the goal of previous approaches \cite{Zhao2017LearningDD}, but rather to model the full set of possible responses. Then, the second novelty of our work lies in our method for re-transforming what our model has learned into a one-to-one setup. Given our teacher model which predicts sets of actions, we use knowledge distillation \citep{Hinton2015DistillingTK} to distil the posterior probability distribution into a student model which predicts the gold action given the context, when fine-tuning on the Jericho Walkthroughs dataset. This probability distribution is far richer than learning from only the hard targets of the dataset, and thus allows the student model to benefit from the richer range of actions the teacher model has learned. As shown in Figure \ref{fig:overview}, during pre-training we learn the set of possible actions from a given context, which ranges from poor actions (`eat coal'), to average actions (`west') to high-quality actions (`put coal in furnace'). Then during fine-tuning we specialise the model to only predict the high-quality actions.

We compare our approach to CALM \citep{yao2020calm}---currently a state-of-the-art approach for action prediction in text-based games---which represents the paradigmatic one-to-one method by training GPT-2 \citep{Radford2019LanguageMA} to generate the next action given the context using the ClubFloyd dataset of human transcripts. Figure \ref{fig:overview} demonstrates that selecting the gold action requires choosing a highly contextually relevant action---i.e. `put coal in furnace'---which might not be well represented in the dataset, whereas we found CALM preferred the more common output `west'.

Although we use text-based games as the testbed for our approach, our method could be extended to generative dialogue tasks---e.g. Reddit \citep{dialogpt}---if paired with a suitable generative model. Many social media-based dialogue datasets often form a tree-like structure, in which there are many comments for any given post \citep{Baumgartner2020ThePR}. This format would naturally allow for the introduction of a latent variable to first learn to predict all of the comments, then to learn to select which comment is best.

In summary, our key contributions are: (1) a novel discrete latent variable neural retrieval architecture, (2) a novel distillation algorithm for disentangling one-to-many and one-to-one tasks; (3) novel quantitative analysis of a dataset of human transcripts. Empirical results show up to 49\% empirical improvement over the current state-of-the-art model on recall@1 on the Jericho Walkthroughs test set. We make our code available for reproducibility.\footnote{\url{https://github.com/BenjaminTowle/PASA}}

\section{Related Work}
Our paper builds upon previous work on applying language priors to text-based games and also work more generally in NLP on one-to-many tasks, as well as the work on knowledge distillation, which is a key technique in our approach.

\paragraph{Language Priors in Text-base Games}
Several previous works have explored learning linguistic priors for text-based games. The overall idea is that instead of relying on reinforcement learning to learn how to play the games through repeated interactions with the same game \citep{He2016DeepRL, Ammanabrolu2019PlayingTG, kga2c, mprc}, the model can rely on general knowledge it has learned about language to give a good sense for which actions are possible and/or may be useful. \citet{blindfolded} explore the use of intrinsic motivation \citep{Pathak2017CuriosityDrivenEB}---i.e. additional loss functions encouraging the model to explore more---to obtain more semantically relevant representations. \citet{lave} use GloVE word embeddings \cite{pennington-etal-2014-glove} for discovering new potentially useful unexplored actions.  \citet{qbert} use the ALBERT pre-trained model \citep{Lan2020ALBERTAL} and perform question answering for intrinsic motivation and to build information about the state. \citet{Singh2022PretrainedLM} use the DistilBERT model \citep{Sanh2019DistilBERTAD} as a base for Q-learning. There has also been work towards using text-based games as a testbed for ethical value alignment in agents \cite{Hendrycks2021WhatWJ}. Most relevant to our work is \citet{yao2020calm} who train GPT-2 \citep{Radford2019LanguageMA} to generate the next action given the context using the ClubFloyd dataset of human transcripts; \citet{worldformer} extend this by multi-tasking on world-modeling tasks, using the Jericho Worlds dataset which extends the Jericho Walkthroughs dataset \citep{jerichoworlds}. However, unlike our work, these works have largely focused on generating \textit{valid} actions, rather than the optimal \textit{gold} action.

\paragraph{One-to-Many Modelling in NLP}
The predominant method for one-to-many modelling is to introduce some latent information into the context. Often this is in the form of a learned latent variable: Variational autoencoders have been used for generating language from continuous spaces \citep{Bowman2016GeneratingSF}; Subsequently, they were extended to dialogue models as conditional variational autoencoders to model the diversity of responses \cite{Zhao2017LearningDD}; \citet{Gao2019JointlyOD} jointly optimise over diversity and relevance within the latent space of an RNN; \citet{Bao2020PLATOPD} train a large transformer with a discrete latent variable to produce more interesting outputs. Different from our approach however, the focus of these approaches is largely on generating more diverse outputs, rather than being able to explicitly generate a set of outputs. Thus, it is difficult to steer the outputs of these models towards a particular type of desirable outputs. Other one-to-many methods involve conditioning the outputs on more interpretable information such as the speaker id \cite{Li2016APN}, or consider the high-level intent of the response \cite{HeHe2018DecouplingSA}. These are also highly relevant to our paper, as we also explore such interpretable latent variables, although again our purpose differs in attempting to capture the full set of possible actions.

\paragraph{Knowledge Distillation}
Due to the rapid upscaling of model size in NLP, Knowledge distillation (KD) is a popular technique for encoding the knowledge of a larger teacher model into a smaller student model \citep{Hinton2015DistillingTK}. KD is most typically used for compressing models of similar types, such as generative models \citep{distilgpt2}, or language understanding models \cite{Sanh2019DistilBERTAD}. Also it can be used for transferring knowledge between heterogeneous models, such as from a reader to a retriever in open-domain QA \cite{Izacard2021DistillingKF}, or from a slow re-ranker to a fast bi-encoder in dialogue response selection \citep{Tahami2020DistillingKF}. To the best of our knowledge however, our work is the first to use knowledge distillation to transfer between one-to-many and one-to-one tasks.     

\section{Method}
\begin{figure*}
    \centering
    \includegraphics[scale=0.09]{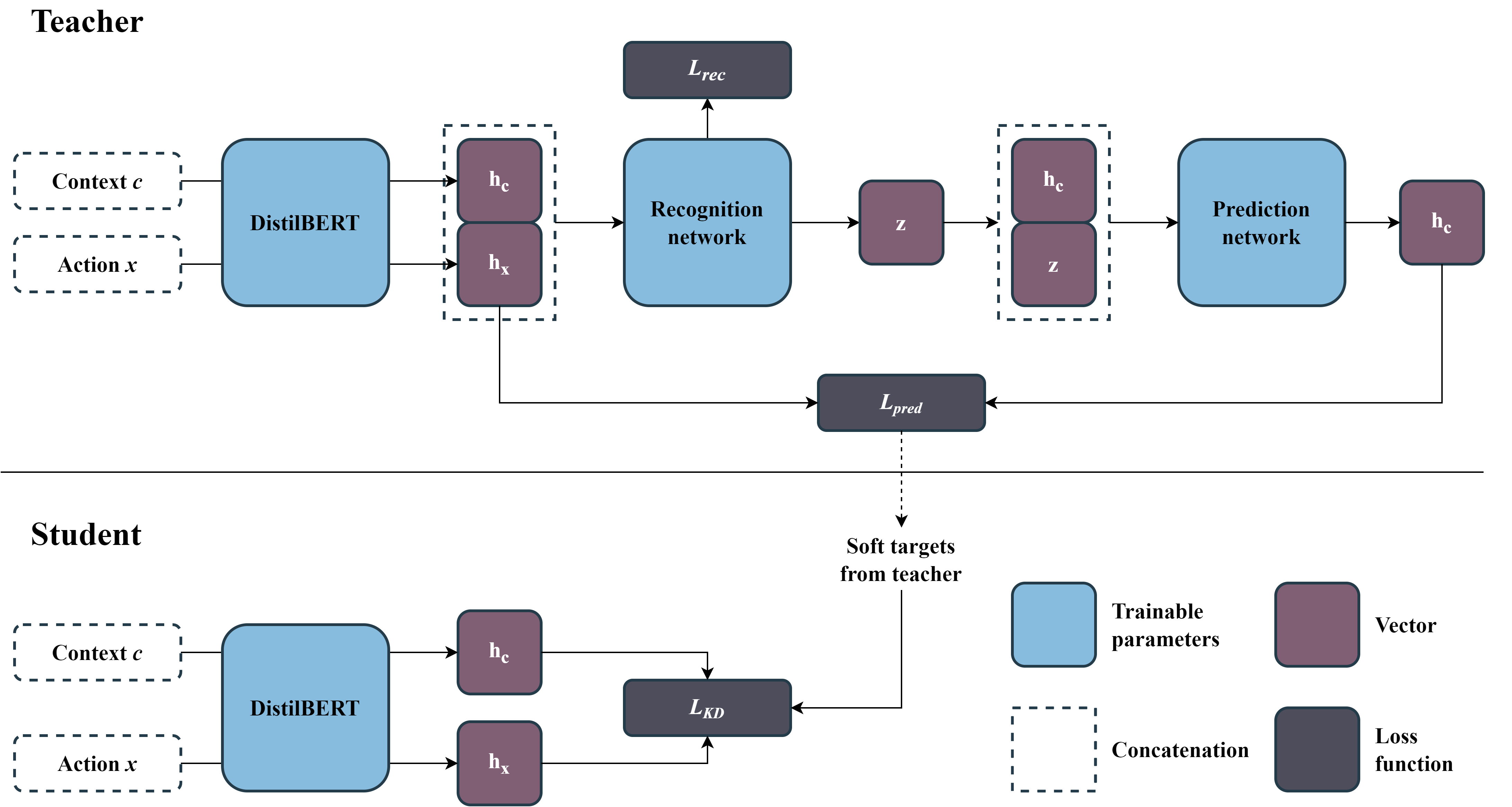}
    \caption{The architecture for \model.}
    \label{fig:model}
\end{figure*}

In this section, we describe our PASA architecture (see Figure \ref{fig:model}), which comprises a teacher model which uses posterior latent variables, and a student model, which is trained via knowledge distillation on the soft targets of the teacher model. Both models are retrieval models, in that they aim to select the gold action, given a subset of actions.

\subsection{Task definition}
We focus on the problem of choosing the gold action in text-based games, as defined by a set of walkthroughs which comprise the sequence of actions required to maximise the score, from the set of valid actions provided by back-end access to the game engine. Formally, we learn to predict the action $a_t$ at time $t$, from the current context $c_t$, which is given as the last $n$ elements from the current trajectory of interwoven actions and observations since the start of the game $<o_0,x_0,...,x_{t-1},o_{t-1}>$.

\subsection{Encoding}
For encoding both contexts and actions, we use the DistilBERT model \citep{Sanh2019DistilBERTAD}, which is a 6-layer 66M parameter bidirectional transformer, trained via knowledge distillation \citep{Hinton2015DistillingTK} on BERT \cite{Devlin2019BERTPO}. We separately encode the context and each action, and aggregate the final hidden states by taking the output vector from the first token, which corresponds to the [CLS] token which is pre-pended to every sentence. We denote $h_c$ and $h_x$ as the vector output from DistilBERT for the context and action respectively.

\subsection{Student}
The student is the simpler of the two models, which scores actions by taking the dot product between the encoded context and action representations and applies a softmax normalisation over this to obtain a probability distribution over valid actions.

\subsection{Teacher}
Open-ended language tasks such as dialogue or summarisation can be framed as one-to-many problems---i.e. there are many acceptable responses/summaries for a particular context. We can represent this by introducing a latent variable $z$, which is a single 1-of-$K$ categorical variable which is also used to condition the action in addition to the context. Intuitively, this allows our model to predict $K$ different actions from a single context, simply by fixing $z$ to different values. To achieve this, we extend our student model with both a recognition network, which learns to predict the latent variable $z$ and a prediction network, which synthesises both the context and chosen value for $z$.

\paragraph{Recognition network.} The recognition network takes as its inputs both the context and action and outputs a posterior probability distribution over $z$. We concatenate the output vectors from each $\langle$context, action$\rangle$ pair and apply two fully connected layers followed by a softmax normalisation:

\[ p(z|x,c) = softmax(W_2(W_1([h_x;h_c]))) \]

Where $[\cdot ; \cdot ]$ represents vector concatenation, $W_1 \in \mathbb{R}^{d \times k}$ and $W_2 \in \mathbb{R}^{2d \times d}$. We then sample from this distribution using the Gumbel-Softmax re-parametrization trick to obtain low-variance gradients \cite{gumbelsoftmax}:

\[ z' = GS(p(z|x,c)) \]

Where $z'$ is a differentiable $K$-dimensional one-hot vector.

\paragraph{Prediction network.}

The prediction network serves to predict the ground-truth action. It takes as its input the context and one-hot encoded latent variable. These are concatenated and put through a fully connected layer where $W_3 \in \mathbb{R}^{(d+k) \times d}$: 

\[ h_{c,z} = W_3([h_c;z']) \]

\subsection{Semantically Aligning the Latent Variable}
In the standard setup for latent variable models the recognition network is reliant on the gradients from end-to-end training in order to learn a good representation for $z$. Intuitively however, we might be able to guide the recognition network by using some heuristics to determine how $z$ should be assigned as an additional loss function $L_{rec}$ which acts as a soft constraint. We present three different approaches for this:

\paragraph{Latent. } This represents the vanilla discrete latent variable approach, in so far as we do not explicitly tie the latent variable to any particular semantics. Instead, to encourage the model to make use of the full capacity of the latent variable, we regularise $z$ by using a uniform prior with batch prior regularisation (BPR) \citep{Zhao2018UnsupervisedDS}. Particularly, we take the approximated marginal distribution $q'(z)$ over the latent variable:

\[ q(z) \approx \frac{1}{N}\sum_n q(z|c_n,x_n) = q'(z) \]

Then, we minimise the Kullback-Leibler (KL) divergence between the uniform prior $p(z)$ and our approximated marginal $q'(z)$.

\[ L_{rec} = \sum_{i} q'(z=z_i)\log\frac{q'(z=z_i)}{p(z=z_i)} \]

Note that this regularisation differs significantly from the more common approach of learning to predict a prior probability given the context and then directly minimising KL-divergence between the prior and posterior \cite{Zhao2017LearningDD}. Since, that approach is known to lead to posterior collapse, whereby the model outputs a fixed distribution for the latent variable regardless of the posterior information in order to trivially reduce the KL-divergence loss term. Despite not explicitly making the latent variable predictable from the context, our model should not learn to ignore the context as the capacity of the latent variable is highly constrained, and therefore the context is still required to accurately predict the action.

\paragraph{Persona.} In a task such as dialogue, an intuitive answer to why different responses can be given to the same context is because there are different people responding, who may each have their own linguistic styles, attitudes and goals. Previous work on persona-grounding has therefore sought to inject speaker information into the context to aid prediction \citep{Li2016APN, personachat}. Inspired by this, we treat each transcript from the pre-training data as a different speaker, where each speaker represents a different value the latent variable $z$ can take. We then add an additional loss term for the recognition network to predict the correct value for $z$ given the context and action, which we train with cross-entropy loss. Note, we do not apply this loss function during fine-tuning as we have no labelled mapping between the speakers in the pre-training data and the creators of the walkthroughs.

\paragraph{Intent.} In task-oriented dialogue, an intent represents a high-level action the user wishes to make, such as a greeting, asking a question, requesting information etc. \citep{Young2013POMDPBasedSS}. Often, these intents can be identified at a coarse level using fairly simple regular expressions \citep{HeHe2018DecouplingSA}. We observe that many actions in text-based games can be hierarchically organised in a similar way. The current state-of-the-art agent for playing text-based games in zero-shot settings is a rules-based model NAIL \citep{Hausknecht2019NAILAG}, which adopts a modular approach identifying four types of action: \textit{navigation} - moving between locations; \textit{examination} - using the examine and look commands to gather feedback about the environment's visuals; \textit{hoarding} - accumulating as many objects as possible; \textit{interacting} - interacting with the environment using objects from the inventory such as fighting a troll with a sword or opening a chest. As presented in Table \ref{tab:regex}, we use regular expressions and the SpaCy part-of-speech tagger \footnote{\url{https://spacy.io/}} to classify actions according to this schema, and add an additional `other' class for anything not matched. Then we train the recognition network to predict the correct class using cross-entropy loss.

\begin{table}[]
\small
    \centering
    \begin{tabular}{lcc}
    \hline
         Intent & Matching Patterns & \# Samples  \\
         \hline
         navigate & north, east, south etc. & 54.6k \\
         examine & examine, look & 39.1k \\
         hoard & take, get, inventory etc. & 18.5k \\
         interact & contains verb and noun & 25.9k \\
         other & -- & 49.6k \\
        \hline
    \end{tabular}
    \caption{Rules for intent detection with number of samples by intent. Matching conditions are tested in the listed order. SpaCy is used for relevant part-of-speech tagging.}
    \label{tab:regex}
\end{table}

\subsection{Training pipeline}
\label{sec:training}
We train our models in four stages: (1) student pre-training, (2) teacher pre-training, (3) teacher fine-tuning, (4) teacher-student knowledge distillation. In each stage we optimise our model with cross-entropy loss to predict the ground truth action, given a selection of negatives:

\[ L_{pred} = - \log \frac{exp(h_c \cdot h_x)}{\sum_j exp(h_{c,j} \cdot h_{x,j})} \]

For pre-training using the ClubFloyd dataset, we use other actions from the batch as negatives, as we do not have access to the backend set of valid actions there. For stages (3) and (4), where we use the Jericho Walkthroughs dataset, we use the valid actions provided by the game engine as the negatives. 

\paragraph{Knowledge distillation.}
In the standard setup for latent variable models it is common to train a prior network which must predict the latent variable given only the context \citep{Zhao2017LearningDD}. This is trained concurrently to the recognition network via reverse KL-divergence, which encourages mode-seeking behaviour from the prior network. However, this prevents our teacher model from learning to predict the full set of possible actions. As shown in \citet{Zhao2018UnsupervisedDS}, there is a conflict between trying to minimise the KL-divergence (important for the one-to-one task) and maximising the information between the actions and latent variable (important for the one-to-many task). Intuitively, to the extent that we can easily predict $z$ from the context $c$, so too is $z$ unlikely to tell us anything new about the action $x$ that we didn't already known from $c$. We therefore approach transitioning from a one-to-many to a one-to-one task through knowledge distillation allowing us to keep the tasks fully disentangled. Particularly, given our discrete latent model fine-tuned on the Jericho Walkthroughs as the teacher $\tau$ and the baseline model pre-trained on Clubfloyd as the student $\sigma$. We then perform distillation on the Jericho Walkthroughs dataset. Specifically, we train the student model using a combination of the soft targets provided by the teacher network $p_\tau(\cdot)$, and the ground-truth hard targets $p_\gamma(\cdot)$ from the dataset itself, weighted by a coefficient $\beta$ (set to 1 during all of our experiments):

\begin{multline*}
L_{KD} = - p_\tau(x|c) * \log p_\sigma(x|c)\\ 
- \beta p_\gamma(x|c) * \log p_\sigma(x|c)
\end{multline*}

Following \citet{Hinton2015DistillingTK}, we apply a temperature of $20$ to the logits for both models before the softmax operation for cross-entropy training on the soft targets only. This is because the teacher typically produces a very sharp probability distribution at a temperature of $1$ - as it has already seen the training data - and therefore the student cannot learn much from the class probabilities.

\subsection{Inference}
During inference, we only use the student model, as the teacher model assumes access to the ground-truth action in order to perform posterior latent variable assignment. Intuitively, we hypothesise that the student model will perform better on the prediction task by having been trained on the richer probability distribution from the teacher model, rather than the hard labels of the dataset.

\section{Experiment}

\subsection{Datasets}

\begin{figure}[t]
    \centering
    \includegraphics[scale=0.4]{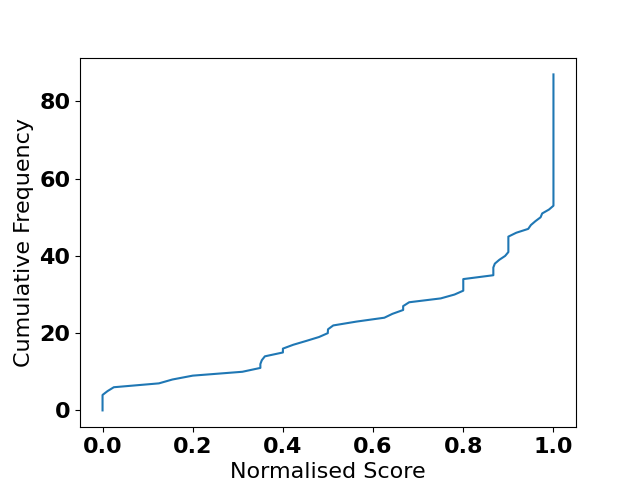}
    \caption{Cumulative frequency graph of normalised player scores from the ClubFloyd dataset.}
    \label{fig:cumulative_frequency}
\end{figure}

\begin{table}[t]
    \small
    \centering
    \begin{tabular}{lll}
    \hline
          & CF & JW  \\
         \hline
         \# Unique Games & 590 & 28 \\
         Vocab size & 39,670 & 9,623 \\
         Vocab size (game avg.) & 2,263 & 1,037 \\
         Train samples & 187,679 & 4,523 \\
         Test samples & 22,947 & 1,626 \\
         Total samples & 210,626 & 6,149 \\
         Avg. steps per reward & 18.2 & 9.7 \\
        \hline
    \end{tabular}
    \caption{Statistics for the ClubFloyd (CF) and Jericho Walkthroughs (JW) datasets.}
    \label{tab:datasets}
\end{table}

\paragraph{The ClubFloyd} \cite{yao2020calm} dataset comprises 426 transcripts of 590 unique games and 210,626 context-action pairs created through human gameplay \footnote{\url{https://github.com/princeton-nlp/calm-textgame}}. We follow the same preprocessing approach as \citet{yao2020calm}: namely (1) we remove transcripts containing games in the Jericho platform, to ensure there is no contamination with the fine-tuning data; (2) We select 10\% of the transcripts to form the validation data, noting that splitting at the transcript level prevents test information from leaking into the training data (e.g. an action appearing in the context trajectory for a train sample being the target for a test sample). 

We apply additional analysis beyond previous work to further understand the differences in action quality for ClubFloyd compared to a walkthrough containing only gold actions. Upon reaching a terminal state, most games will report the player's final score as well as the total possible score. Through pattern matching, we are able to extract final scores for 83 of the transcripts, which we normalise by dividing by the total possible score. We present these in Figure \ref{fig:cumulative_frequency}. As shown, there is significant variation in performance. Additionally, even the top scoring performances are far from being expert demonstrations, as these are only the final scores after players have completed large amounts of exploration, sometimes saving and loading between game states. Some games also report whenever the player's score changes, which we identify using regular expressions in 115 of the transcripts. In Table \ref{tab:datasets} we also report the average steps required per reward (only counting transcripts where at least a single reward is observed). We see that the walkthroughs require approximately half as many steps per reward, further validating the hypothesis that ClubFloyd contains suboptimal actions on average.

\paragraph{The Jericho Walkthroughs} dataset contains 28 walkthroughs for the games supported by the Jericho platform \footnote{\url{https://github.com/microsoft/jericho}}. Although these walkthroughs contain only the desired behaviour---i.e. gold actions---they are limited in number and therefore difficult to generalise from. Following previous work \cite{jerichoworlds}, we split the train/test data by game, ensuring that the model is evaluated on unseen games. This is important to prevent contamination as the context for one sample also includes the ground truth action for another. However, it also adds an increased layer of difficulty as some games include unusual vocabulary, such as the name of a spell, which does not exist outside of the game. Unseen games therefore present a difficult generalisation task which requires significant language understanding. The games evaluated on are classified as either \textit{difficult} or \textit{possible} games, according to \citet{colossal}. This takes into account a variety of factors such as the size of the action space, how unusual the actions are (i.e. non-standard words such as the name of a spell), how sparse the reward is, and the complexity of the tasks required (easier games may reward the player for basic exploration, e.g. `west' `east' etc., while difficult games may require more reasoning such as moving a rug to reveal a trap door).

\begin{table*}[t]
\small
\centering
\begin{tabular}{l|l|ll|lll|l}
\hline
& & \multicolumn{2}{c|}{CALM} & \multicolumn{3}{c|}{PASA} & Ablation  \\
& \# valid acts & PT & FT & Persona & Latent & Intent & Student-only \\
\hline
\multicolumn{1}{l}{\textit{Difficult Games}}   \\
\hline
zork1     & 16.9 & 30.1 & 33.1 & \textbf{53.0} & 48.2 & 52.8 & 51.5  \\
deephome  & 20.4 & 44.6 & 43.1 & 5.5 & \textbf{40.1} & 36.1 & 14.7 \\
ludicorp  & 15.5 & 43.7 & 54.1 & \textbf{70.6} & 69.2 & \textbf{70.6} & 66.2 \\
balances  & 24.2 & 29.5 & 28.7 & 54.9 & \textbf{68.9} & 64.8 & 49.2 \\
\hline
\multicolumn{1}{l}{\textit{Possible Games}}    \\
\hline
pentari   & 5.3 & 34.7 & 36.7  & \textbf{44.9} & 40.8 & \textbf{44.9} & 40.8 \\
detective & 8.1 & 60.8 & 47.1  & 52.9 & \textbf{58.8} & 56.9 & 52.9 \\
library   & 8.6 & 26.9 & 26.9  & \textbf{34.6} & 28.8 & 30.8 & 32.7 \\
temple    & 16.1 & 31.5 & 49.2  & 74.6 & 73.5 & \textbf{79.6} & 70.7 \\
ztuu      & 34.8 & 29.8 & 33.3  & 21.4 & \textbf{29.8} & 26.2 & 27.4 \\
\hline
overall   & 17.8 & 37.1 & 41.6 & 47.5$\dagger$ & 54.2$\dagger$ & \textbf{55.3$\dagger\ddagger$} & 47.2 \\
\hline
\end{tabular}
\caption{Recall@k performance on the Jericho Walkthroughs test set. Games are classified as either `possible' or `difficult', using the classification from \citet{colossal}. $\dagger$/$\ddagger$ = statistically significant on t-test with \textit{p}-value < 0.01 compared to CALM-FT and Student-only ablation respectively.}
\label{tab:main_results}
\end{table*}

\subsection{Experimental Setup}
We train our models using the Adam optimizer \citep{Kingma2015AdamAM}. We train for a single epoch during pre-training and 3 epochs during fine-tuning with a learning rate of 5e-5. We use a batch size of 8. All of our methods use the DistilBERT model as their base, which provides a good trade-off between size and accuracy being 40\% smaller than BERT, while retaining 97\% of the performance across NLP tasks. We truncate contexts to the last three elements following \citet{yao2020calm} such that $c_t = <o_{t-1},x_{t-1},o{t}>$. During tokenization, we truncate the contexts and actions to a maximum of 128 and 8 tokens respectively. Following previous approaches \citep{jerichoworlds, worldformer}, we evaluate on the Jericho Walkthroughs test set, while our model is trained first on the ClubFloyd dataset, then fine-tuned on the Jericho Walkthroughs train set, following the procedure outlined in Sec \ref{sec:training}. We evaluate using recall@1, which measures the model's ability to select the gold action from the pool of valid actions, which is a standard evaluation metric in retrieval tasks \citep{Ouz2021DomainmatchedPT}.

\subsection{Baseline}
For our baseline we use the CALM model \citep{yao2020calm} \footnote{\url{https://github.com/princeton-nlp/calm-textgame}}\footnote{Another possible baseline candidate would have been the Worldformer \cite{worldformer}, however the code is not released and our evaluation task is different from the one they report on, therefore comparison is not possible.}, which is a state-of-the-art approach to action prediction in text-based games and represents the paradigmatic one-to-one method by training GPT-2 \citep{Radford2019LanguageMA} to generate the next action given the context using the ClubFloyd dataset of human transcripts. Following their implementation, we score actions by summing over the token-wise log-probabilities for each action. Additionally, we also further fine-tuned this model directly on the Jericho Walkthroughs dataset, to make for fair comparison with our approach. We refer to these approaches as CALM-PT and CALM-FT respectively. 

\subsection{Results}
\label{sec:analysis}
We report recall@1 performance on Jericho Walkthroughs in Table \ref{tab:main_results}, with individual performance for each game, split according to the difficulty classification provided by \citet{colossal}. Although some of the factors in assigning difficulty are more relevant from a reinforcement learning perspective, e.g. the sparsity of the reward, many of them also carry across to our more supervised learning setting, such as the number of valid actions and the size and richness of the vocabulary.

Overall, our best version of PASA (intent) achieves a 49\% improvement versus CALM-PT \citep{yao2020calm}. Our approach improves significantly on all difficult games besides \textsc{Deephome}. The difficult games typically require more complex actions to be selected, which require more sophisticated interactions with the environment, such as knowing that the coal should be put in the furnace after it had been opened in Figure \ref{fig:overview}, thus demonstrating the benefit of our approach. In the outstanding case of \textsc{Deephome}, we found this was due to the model overpredicting highly out-of-distribution actions (see Section \ref{error_analysis}). The generative model (CALM) however assigned much lower scores to out-of-distribution actions, due to their tendency towards more generic actions \cite{li-etal-2016-diversity}. Our approach also surpasses in most of the possible games, namely \textsc{Pentari}, \textsc{Temple} and \textsc{Library}, though by a smaller margin. This is to be expected, as most of the actions there consist of relatively simple actions such as navigation `west', `north' etc. which is likely to be well-represented in the pre-training data. 

When comparing the different choices of latent variables, we find the best results come from the intent-based method, suggesting the chosen intents map well onto the text-based game structure. However, the latent approach also followed in a close second, supporting the idea that our approach could be generalised to other NLP tasks where an intent-based method might not be available. The persona-based method was significantly worse, although still better than the CALM approach. This may be because the personas learned in the ClubFloyd dataset did not map as well onto the Jericho Walkthroughs. Overall however, all our approaches were statistically significantly better than the CALM approaches on a t-test with \textit{p}-value < 0.01.

We found some, though limited, benefit to fine-tuning CALM on the ClubFloyd dataset, suggesting the primary source of the benefit is from our architectural decision to disentangle the one-to-many relation, rather than the addition of the Jericho Walkthroughs dataset. This is further supported by our ablation---presented in Table \ref{tab:main_results}---which compares PASA to the student model trained solely on the hard labels from the datasets without access to the soft labels from the teacher. The purpose of this was to isolate the performance gains due to the posterior distillation which is a novelty of our approach. Again, both our latent and intent methods were statistically significantly better than this on a t-test with \textit{p}-value < 0.01. 

\subsection{Error Analysis}
\label{error_analysis}
We inspected the model predictions to further understand the driving factors behind the significant performance differences between different versions of PASA for some games (e.g. \textsc{Deephome} and \textsc{Balances}). We found the persona model overly predicted the out-of-distribution actions `say manaz' (the name of a spell) and `spells' (which causes the game to tell the player which spells they know). Both of these are available as actions for the majority of each respective game, thereby amplifying the impact of the error. We hypothesise the reason the persona model specifically suffers from this problem may be because the latent variable it uses has more representational capacity, being a 1-of-375 categorical variable corresponding to the 375 users in the training data (versus 1-of-5 and 1-of-8 for the other two approaches). Thus, it may have overfitted more to the training data.

\section{Conclusion and Future Work}
Overall we have demonstrated an approach that disentangles one-to-many and one-to-one prediction by first training a discrete latent variable model with posterior-guided latent variable selection and then distilling this into a student model. Results showed up to 49\% empirical improvement over the current state-of-the-art approach. 
 
Our proposed approach currently assumes we have access to a ground-truth set of `valid actions'. This holds in the case of many dialogue datasets from social media \citep{Baumgartner2020ThePR}, as there are often many replies (sc.~valid actions) to a given post. In other tasks however such as summarisation or translation we may only have a single exemplum to learn from, and therefore additional techniques would be required to utilise PASA in those cases.

We chose text-based games as a starting point in this paper  because of the availability of the set of possible actions in the Jericho dataset, which allowed us to isolate the task of selecting the best action in evaluation. However, expanding these techniques to other NLP domains such as dialogue would be a key next step for future work. As an example, our approach could be extended to Reddit \citep{Baumgartner2020ThePR} by treating the number of upvotes to a given comment as the reward, and then treating all of the comments to a particular post as the set of possible actions. Then, the evaluation could measure the model's ability to select the comment with the most upvotes given all the comments to a particular post. 

\bibliography{main}
\bibliographystyle{acl_natbib}

\appendix

\section{Hyperparameters}
In Table \ref{tab:hyperparameters} we report the hyperparameters used during our experiment. Note, we did not search over hyperparameters but relied upon recommendations from previous works \citep{Devlin2019BERTPO}. All of our models were trained using the PyTorch framework \citep{pytorch}.
\begin{table}[]
    \centering
    \begin{tabular}{c|c|c}
    \hline
         & CF & JW  \\
         \hline
         Learning-rate & 5e-5 & 5e-5 \\
         Epochs & 1 & 3 \\
         Batch size & 8 & 8 \\
         Max. context tokens & 128 & 128 \\
         Max. action tokens & 8 & 8 \\
         \hline
    \end{tabular}
    \caption{Hyperparameters for training on the ClubFloyd (CF) and Jericho Walkthroughs (JW) datasets.}
    \label{tab:hyperparameters}
\end{table}
\label{sec:appendix}

\section{Validation}
We split the transcripts from ClubFloyd randomly into a 90/10 train/validation split. Table \ref{tab:validation} presents the validation results for our PASA model. We report on recall@1, which is given as the proportion of the time that the model selects the correct action from 9 other candidates randomly selected from the rest of the validation data (unlike Jericho Walkthroughs, we do not have access to the set of valid actions). Note that the model has posterior access to the ground truth for selecting the latent variable, and so is expected to achieve a high result. We also present the student model for comparison, which does not have access to the ground truth. The difference in performance therefore is an indication of how much information each latent type is able to encode. We find the intent model is the best by a significant margin, which was later supported by our main findings in the Jericho Walkthroughs dataset.

\begin{table}[t]
    \centering
    \begin{tabular}{c|c}
    \hline
        Model & R@1 \\
        \hline
        Student & 53.3 \\
        \hline
        PASA-Latent & 57.8 \\
        PASA-Persona & 59.4 \\
        PASA-Intent & 65.0 \\
        \hline
    \end{tabular}
    \caption{Recall@1 on the ClubFloyd valiation set for PASA and the student model.}
    \label{tab:validation}
\end{table}

\section{Runtime}
Our models are all very efficient to run, as (a) they rely upon using a model that has already been pre-trained, (b) they use the DistilBERT model, which is 40\% smaller than BERT. Overall, no model took more than 1 hour to train on an NVIDIA P100 GPU.

\end{document}